\documentclass[10pt, a4paper, conference]{IEEEtran}
\ifCLASSINFOpdf
  \usepackage[pdftex]{graphicx}
\else
\fi
\usepackage[cmex10]{amsmath}
\usepackage{bbold}

\begin{document}
%
\title{Implicitly Constrained Semi-Supervised Linear Discriminant Analysis}



\author{\IEEEauthorblockN{Jesse H. Krijthe\IEEEauthorrefmark{1}\IEEEauthorrefmark{3},
Marco Loog\IEEEauthorrefmark{2}\IEEEauthorrefmark{3}}
\IEEEauthorblockA{jkrijthe@gmail.com, m.loog@tudelft.nl\\ \IEEEauthorrefmark{1}Pattern Recognition Laboratory, Delft University of Technology, The
Netherlands}
\IEEEauthorblockA{\IEEEauthorrefmark{2}The Image Group, Department of Computer Science, University of Copenhagen, Denmark}
\IEEEauthorblockA{\IEEEauthorrefmark{3}Department of Molecular Epidemiology, Leiden University Medical Center, The Netherlands}}

%


\maketitle

\begin{abstract}
Semi-supervised learning is an important and active topic of research in pattern recognition. For classification using linear discriminant analysis specifically, several semi-supervised variants have been proposed. Using any one of these methods is not guaranteed to outperform the supervised classifier which does not take the additional unlabeled data into account. In this work we compare traditional Expectation Maximization type approaches for semi-supervised linear discriminant analysis with approaches based on intrinsic constraints and propose a new principled approach for semi-supervised linear discriminant analysis, using so-called implicit constraints. We explore the relationships between these methods and consider the question if and in what sense we can expect improvement in performance over the supervised procedure. The constraint based approaches are more robust to misspecification of the model, and may outperform alternatives that make more assumptions on the data in terms of the log-likelihood of unseen objects.
\end{abstract}


%
\IEEEpeerreviewmaketitle

\section{Introduction}
In many real-world pattern recognition tasks, obtaining labeled examples to train classification algorithms is much more expensive than obtaining unlabeled examples. These tasks include document and image classification \cite{Nigam2000} where unlabeled objects can easily be downloaded from the web, part of speech tagging \cite{Elworthy1994}, protein function prediction \cite{Weston2005} and many others. Using unlabeled data to improve the training of a classification procedure, however, requires semi-supervised variants of supervised classifiers to make use of this additional unlabeled data. Research into semi-supervised learning has therefore seen an increasing amount of interest in the last decade \cite{Chapelle2006}.

In supervised learning adding additional labeled training data improves performance for most classification routines. This does not generally hold for semi-supervised learning \cite{Cozman2006}. Adding additional unlabeled data may actually deteriorate classification performance. This can happen when the underlying assumptions of the model do not hold. In effect, disregarding the unlabeled data can lead to a better solution.

In this work we consider linear discriminant analysis (LDA) applied to classification. Several semi-supervised adaptations of this supervised procedure have been proposed. These approaches may suffer from the problem that additional unlabeled data degrade performance. To counter this problem, \cite{Loog2014a} introduced moment constrained LDA, which offers a more robust type of semi-supervised LDA. The recently introduced idea of implicitly constrained estimation \cite{Krijthe2013}, is another method that relies on constraints given by the unlabeled data. We compare these two approaches to other semi-supervised methods, in particular, expectation maximization and self-learning, and empirically study in what sense we can expect improvement by employing any of these semi-supervised methods.

The contributions of this work are the following:

\begin{itemize}
  \item Introduce a new, principled approach to semi-supervised LDA: implicitly constrained LDA
  \item Offer a comparison of semi-supervised versions of linear discriminant analysis
  \item Explore ways in which we can expect these semi-supervised methods to offer improvements over the supervised variant, in particular in terms of the log likelihood
\end{itemize}

The rest of this paper is organized as follows. After discussing related work, we introduce several approaches to semi-supervised linear discriminant analysis. These methods are then compared on an illustrative toy problem and in an empirical study using several benchmark datasets. We end with a discussion of the results and conclude.

\section{Related Work}
Some of the earliest work on semi-supervised learning was done by \cite{McLachlan1975,Taylor1977} who studied the self-learning approach applied to linear discriminant analysis. This has later also been referred to as Yarowsky's algorithm \cite{Yarowsky1995}. This approach is closely related to Expectation Maximization \cite{Abney2004}, where, in a generative model, the unknown labels are integrated out of the likelihood function and the resulting marginal likelihood is maximized \cite{Dempster1977}. More recent work on discriminative semi-supervised learning has focussed on introducing assumptions that relate unlabeled data to the labeled objects \cite{Chapelle2006}. These assumptions usually take the form of either a manifold assumption \cite{Zhu2003}, encoding that labels change smoothly in a low-dimensional manifold, or a low-density class separation assumption used in, for instance, transductive support vector machines \cite{Bennett1998,Joachims1999} and entropy regularization \cite{Grandvalet2005}. 

Work on semi-supervised LDA has tried to incorporate unlabeled data by leveraging the increase in accuracy of estimators of quantities that do not rely on labels. An approach relying on the more accurate estimate of the total covariance matrix of both labeled and unlabeled objects is taken for dimensionality reduction in Normalized LDA, proposed by \cite{Fan2008} and similar work by \cite{Cai2007}. In addition to this covariance matrix, \cite{Loog2014a} also include the more accurate estimate of the overal mean of the data and propose two solutions to solve a subsequent optimization problem. Building on these results, in \cite{Krijthe2013} we introduced implicitly constrained least squares classification, a semi-supervised adaptation of least squares classification. Since this procedure proved both theoretically and practically successful for a discriminative classifier, here we consider whether the idea of implicitly constrained semi-supervised learning can be extended to generative classifiers such as LDA.

\section{Methods}
We will first introduce linear discriminant analysis as a supervised classification algorithm and discuss different semi-supervised procedures. We will consider 2-class classification problems, where we are given an $N_l \times d$ design matrix $\mathbf{X}$, where $N_l$ is the number of labeled objects and $d$ is the number of features. For these observations we are given a label vector $\mathbf{y}=\{0,1\}^{N_l}$. Additionally, in the semi-supervised setting we have an $N_u \times d$ design matrix $\mathbf{X}_u$ without a corresponding $\mathbf{y}_u$ for the unlabeled observations.

\subsection{Supervised LDA}
In supervised linear discriminant analysis, we model the 2 classes as having multivariate normal distributions with the same covariance matrix $\boldsymbol{\Sigma}$ and differing means $\boldsymbol{\mu}_1$ and $\boldsymbol{\mu}_2$. To estimate the parameters of this model, we maximize the likelihood, or, equivalently, the log likelihood function:
\begin{align}
\label{eq:lda}
L(\theta|\mathbf{X},\mathbf{y})= \sum_{i=1}^{N_l} & y_i \log(\pi_1 \mathcal{N}(\mathbf{x}_i|\boldsymbol{\mu}_1,\mathbf{\Sigma})) \nonumber \\
& +(1-y_i) \log(\pi_2 \mathcal{N}(\mathbf{x}_i|\boldsymbol{\mu}_2,\mathbf{\Sigma}))
\end{align}
where $\theta=\left( \pi_1,\pi_2, \boldsymbol{\mu}_1,\boldsymbol{\mu}_2,\mathbf{\Sigma} \right)$, $\mathcal{N}(\mathbf{x}_i|\boldsymbol{\mu},\mathbf{\Sigma})$ denotes the density of a multivariate normal distribution with mean $\boldsymbol{\mu}$ and covariance $\mathbf{\Sigma}$ evaluated at $\mathbf{x}_i$ and $\pi_c$ denotes the prior probability for class $c$. The closed form solution to this maximization is given by the estimators:
\begin{align}
\label{eq:ldasolution}
\hat{\pi_1} &= \frac{\sum_{i=1}^{N_l} y_i}{N_l}, \hspace{20px} \hat{\pi_2} = \frac{\sum_{i=1}^{N_l} (1-y_i)}{N_l}  \nonumber \\
\hat{\boldsymbol{\mu}}_1 &= \frac{\sum_{i=1}^{N_l} y_i \mathbf{x}_i}{\sum_{i=1}^{N_l} y_i}, \hspace{20px} \hat{\boldsymbol{\mu}}_2 = \frac{\sum_{i=1}^{N_l} (1-y_i) \mathbf{x}_i}{\sum_{i=1}^{N_l} (1-y_i)}  \nonumber  \\
\hat{\mathbf{\Sigma}} &= \frac{1}{N_l} \sum_{i=1}^{N_l} y_i (\mathbf{x}_i-\boldsymbol{\mu}_1) (\mathbf{x}_i-\boldsymbol{\mu}_1)^T  \nonumber \\
& \hspace{20px} + (1-y_i) (\mathbf{x}_i-\boldsymbol{\mu}_2) (\mathbf{x}_i-\boldsymbol{\mu}_2)^T 
\end{align}
Where the maximum likelihood estimator $\hat{\mathbf{\Sigma}}$ is a biased estimator for the covariance matrix. Given a set of labeled objects $(\mathbf{X},\mathbf{y})$, we can estimate these parameters and find the posterior for a new object $\mathbf{x}$ using:
\begin{equation}
p(c=1|\mathbf{x})=\frac{\pi_1 \mathcal{N}(\mathbf{x}|\hat{\boldsymbol{\mu}}_1,\hat{\mathbf{\Sigma}})}{\sum_{c=1}^{2} \pi_c \mathcal{N}(\mathbf{x}|\hat{\boldsymbol{\mu}}_c,\hat{\mathbf{\Sigma}})}
\end{equation}
This posterior distribution can be employed for classification by assigning objects to the class for which its posterior is highest. We now consider several adaptations of this classification procedure to the  semi-supervised setting.

\subsection{Self-Learning LDA (SLLDA)}
A common and straightforward adaptation of any supervised learning algorithm to the semi-supervised setting is self-learning, also known as bootstrapping or Yarowsky's algorithm \cite{McLachlan1975,Yarowsky1995}. Starting out with a classifier trained on the labeled data only, labels are  predicted for the unlabeled objects. These objects, with their imputed labels are then used in the next iteration to retrain the classifier. This new classifier is now used to relabel the unlabeled objects. This is done until the predicted labels on the unlabeled objects converge. \cite{Abney2004} studies the underlying loss that this procedure minimizes and proves its convergence.

\subsection{Expectation Maximization LDA (EMLDA)}
Assuming the mixture model of Equation \eqref{eq:lda} and treating the unobserved labels $\mathbf{y}_u$ as latent variables, a possible adaptation of this model is to add a term for the unlabeled data to the objective function and to integrate out the unknown labels, $\mathbf{y}_u$, to find the marginal likelihood:
\begin{align}
\l(\theta|\mathbf{X},\mathbf{y},\mathbf{X}_u)=\prod_{i=1}^{N_l}\left(\pi_1 \mathcal{N}(\mathbf{x}_i|\boldsymbol{\mu}_1,\mathbf{\Sigma})\right)^{y_i} \left(\pi_2 \mathcal{N}(\mathbf{x}_i|\boldsymbol{\mu}_2,\mathbf{\Sigma})\right)^{1-y_i}  \nonumber \\ 
\times \prod_{i=1}^{N_u} \sum_{c=1}^2 \pi_c \mathcal{N}(\mathbf{x}_i|\boldsymbol{\mu}_c,\Sigma)
\end{align}
Maximizing this marginal likelihood, or equivalently, the log of this function is harder then the supervised objective in Equation \eqref{eq:lda}, since the expression contains a log over a sum. However, we can solve this optimization problem using the well-known expectation maximization (EM) algorithm \cite{Dempster1977, Nigam2000}. In EM, the log over the sum is bounded from below through Jensen's inequality. In the M step of the algorithm, we maximize this bound by updating the parameters using the imputed labels obtained in the E step. In practice, the M step consists of the same update as in Equation \eqref{eq:ldasolution}, where the sum is no longer over the labeled objects but also the unlabeled objects using the imputed posteriors, or responsibilities, from the E step. In the E step the lower bound is made tight by updating the imputed labels using the posterior under the new parameter estimates. This is done until convergence. 
In effect this procedure is very similar to self-learning, where instead of hard labels, a probability over labelings is used. Both self-learning and EM suffer from the problem of wrongly imputed labels that can reinforce their wrongly imputed values because the parameters are updated as if they were the true labels.

\subsection{Moment Constrained LDA (MCLDA)}
An alternative to the EM-like approaches like EMLDA and SLLDA was proposed by \cite{Loog2010} in the form of moment constrained parameter estimation. The main idea is that there are certain constraints that link parameters that are calculated using feature values alone, with parameters which require the labels. In the case of LDA \cite{Loog2014a}, for instance, the overal mean of the data is linked to the means of the two classes through:
\begin{equation}
\label{eq:constraintmean}
\boldsymbol{\mu}_t=\pi_1 \boldsymbol{\mu}_1 + \pi_2 \boldsymbol{\mu}_2
\end{equation}
Were $\boldsymbol{\mu}_t$ is the overal mean on all the data and therefore does not depend on the labels.
The total covariance matrix $\mathbf{\Sigma}_t$ is linked to the within-class covariance matrix $\mathbf{\Sigma}$ and between-class covariance matrix $\mathbf{\Sigma}_b$, the covariance matrix of the means. Only the latter two rely on the labels:
\begin{equation}
\label{eq:constraintcovariance}
\mathbf{\Sigma}_t=\mathbf{\Sigma} + \mathbf{\Sigma}_b
\end{equation}
Recognizing that the unlabeled data allow us to more accurately estimate the parameters in these constraints that do not rely on the labels, \cite{Loog2014a} points out that this more accurate estimate will generally violate the constraints, meaning the other label-dependent estimates should be updated accordingly.

An ad hoc way to update the parameters based on these more accurate estimates \cite{Loog2014a} leads to the following updated moment constrained estimators:
\begin{align}
\hat{\boldsymbol{\mu}}_c^{MC} & =\hat{\boldsymbol{\mu}}_c - \sum_{j=1}^{2} \hat{\pi}_j \hat{\boldsymbol{\mu}}_j-\hat{\boldsymbol{\mu}}_t \\
\hat{\mathbf{\Sigma}}^{MC} &= \hat{\mathbf{\Theta}}^{\frac{1}{2}} \hat{\mathbf{\Sigma}}_t^{\frac{1}{2}} \hat{\mathbf{\Sigma}} \hat{\mathbf{\Sigma}}_t^{\frac{1}{2}} \hat{\mathbf{\Theta}}^{\frac{1}{2}}
\end{align}
where $\hat{\boldsymbol{\mu}}_t$ and $\hat{\mathbf{\Theta}}$ are the overal mean and overal covariance estimated on all labeled and unlabeled data, while $\hat{\mathbf{\Sigma}}_t$ is the overal covariance estimated on the labeled data alone.

Alternatively and slightly more formally, \cite{Loog2012b} forces the constraints to be satisfied by maximizing the likelihood on the labeled objects under the constraints in Equations \eqref{eq:constraintmean} and \eqref{eq:constraintcovariance}. This leads to a non-convex objective function that can be solved numerically. In this work we use the simpler ad hoc constraints.

\subsection{Implicitly Constrained LDA (ICLDA)}
The former approach requires the identification of specific constraints. Ideally, we would like these constraints to emerge implicitly from a choice of supervised learner and a given set of unlabeled objects. Implicitly constrained semi-supervised learning attempts to do just that. The underlying intuition is that if we could enumerate all possible $2^{N_u}$ labelings, and train the corresponding classifiers, the classifier based on the true but unknown labels is in this set. This classifier would generally outperform the supervised classifier. Two problems arise:

\begin{enumerate}
\item How do we find a classifier in this set that is close to the one based on the true but unknown labels?
\item How do we efficiently traverse this enormous set of possible labelings without having to enumerate them all?
\end{enumerate}
As for the first problem: a safe way to know how well a solution performs in terms of our supervised objective is to estimate its performance using the labeled objects. We therefore propose the following objective:

\begin{equation}
\label{eq:iclda}
\operatorname*{arg\,max}_{\left( \pi_1,\pi_2, \boldsymbol{\mu}_1,\boldsymbol{\mu}_2,\mathbf{\Sigma}\right) \in \mathcal{C}_\theta} L(\pi_1,\pi_2, \boldsymbol{\mu}_1,\boldsymbol{\mu}_2,\mathbf{\Sigma}|\mathbf{X},\mathbf{y})
\end{equation}
where
\begin{align}
\mathcal{C}_{\boldsymbol{\theta}} = \left\{ \operatorname*{arg\,max} L(\pi_1,\pi_2, \boldsymbol{\mu}_1,\boldsymbol{\mu}_2,\mathbf{\Sigma}| \mathbf{X}_e, \mathbf{y}_e) : \mathbf{y}_u \in [0,1]^{N_u} \right\} \nonumber
\end{align}
and $\mathbf{X}_e=[\mathbf{X}^T \mathbf{X}_u]^T, \mathbf{y}_e=[\mathbf{y}^T \mathbf{y}_u^T]^T$ are the design matrix and class vector extended with the unlabeled data. This can be interpreted as optimizing the same objective function as supervised LDA, with the additional constraint that the solution has to attainable by a particular assignment of responsibilities (partial assignments to classes) for the unlabeled objects.

As for the second problem: since, for a given imputed labeling, we have a closed form solution for the parameters, the gradient of the supervised loss \eqref{eq:iclda} with respect to the responsibilities $\mathbf{y}_u$ can be found using
\begin{equation}
\frac{\partial L(\theta|\mathbf{X},\mathbf{y})}{\partial \mathbf{y_u}} = \frac{\partial L(\theta|\mathbf{X},\mathbf{y})}{\partial \theta} \frac{\partial \phi(\mathbf{y}_u)}{\partial \mathbf{y}_u} 
\end{equation}
where $\phi(\mathbf{y}_u)=\theta$ is the function that has as input a particular labeling of the points, and outputs the parameters $\theta=\left(\pi_1,\pi_2, \boldsymbol{\mu}_1,\boldsymbol{\mu}_2,\mathbf{\Sigma}\right)$, similar to Equation \eqref{eq:ldasolution}.

This can be used to efficiently move through the set of solutions using a simple gradient ascent procedure that takes into account the $[0,1]$ bounds on the responsibilities.

\section{Experimental setup and results}
We present simulations on an illustrative toy dataset and a set of benchmark datasets. Other than the classifiers covered in the previous section, we also include the LDA classifier trained using all labels of the unlabeled data (LDAoracle) as an upper bound on the performance of any semi-supervised procedure. The experiments can be reproduced using code from the authors' website. 

\subsection{Toy problems}
\begin{figure*}[!t]
\centering
\includegraphics[scale=0.45]{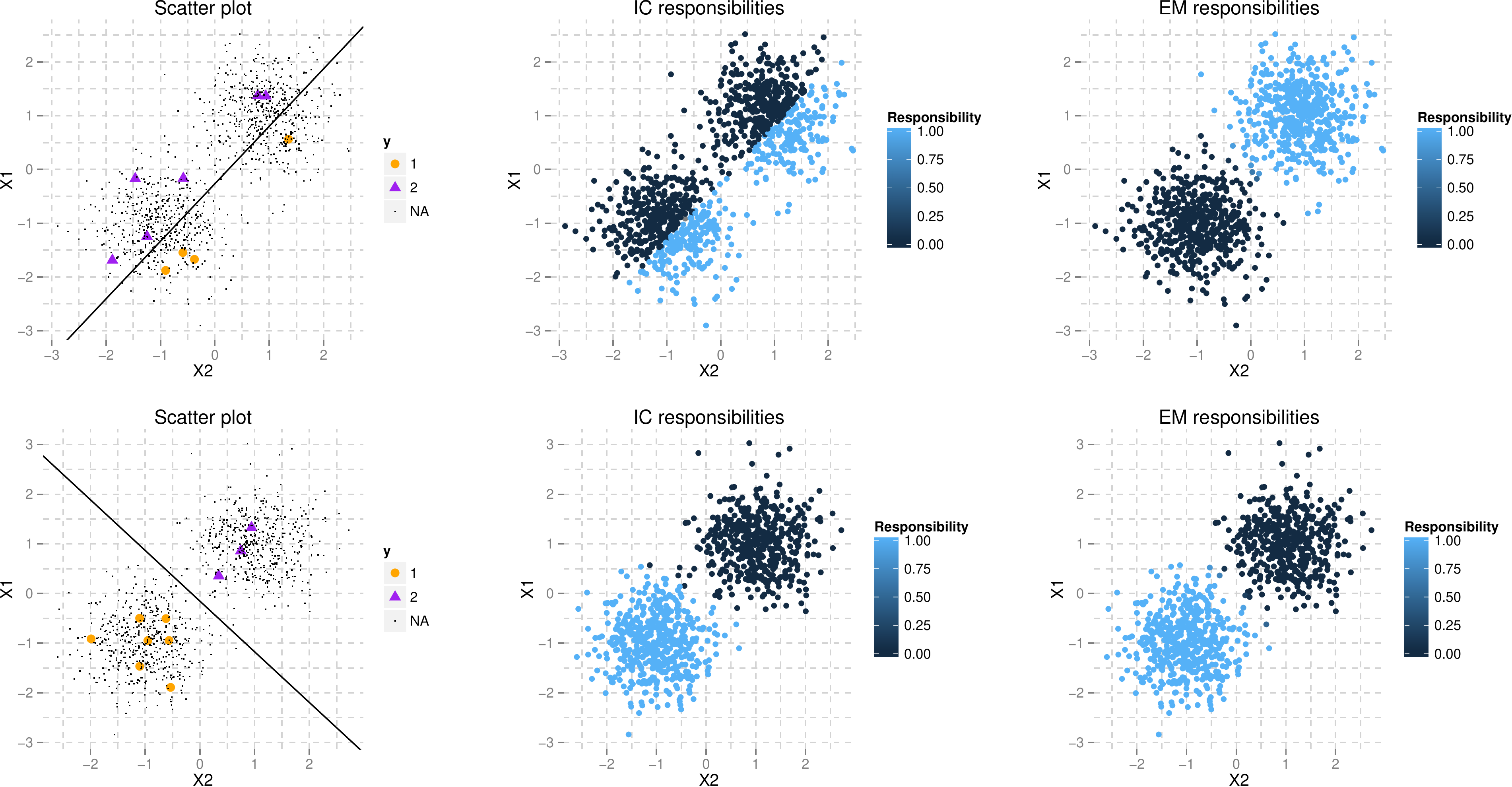}
\caption{Behaviour on the two-class two dimensional gaussian datasets, with 10 labeled objects and 990 unlabeled objects. The first row shows the scatterplot and the trained responsibilities for respectively ICLDA and EMLDA on a dataset where the decision boundary does not adhere to the assumptions of EMLDA. The second row shows the results when the decision boundary is in between the two Gaussian classes. The black line indicates the decision boundary of a supervised learner trained using only the labeled data. Note that in the first row, the responsibilities of EM are very different from the true labels, while IC is not as sensitive to this problem.}
\label{fig:toyplots}
\end{figure*}

To illustrate the behaviour of ICLDA when compared to EMLDA we consider two toy datasets. In both cases we have two multivariate normal distributions centered at respectively $\boldsymbol{\mu}_1=[1,1]^T$ and $\boldsymbol{\mu}_2=[-1,-1]^T$ and equal covariance $\mathbf{\Sigma}=0.6 \mathbf{1}$, with $\mathbf{1}$ the $2 \times 2$ identity matrix. An example is given in Figure \ref{fig:toyplots}. In the bottom row, these two gaussians correspond to the different classes. In the top row, we consider the case where the decision boundary is actually perpendicular to the boundary in the other setting. This means that the bottom row corresponds exactly to the assumptions of EM, while this is not the case in the top row. Figure \ref{fig:toyplots} illustrates what happens in a particular sample from this problem were we draw 10 labeled and 990 unlabeled objects. When the assumption does not hold, EMLDA forces the decision boundary to fall between the two gaussian clusters leading to a much worse solution then the supervised LDA based on only a few labeled examples. The ICLDA solution does not deviate from the correct boundary by much. When the assumptions do hold, EMLDA finds the correct boundary, as expected, while ICLDA only does minor adjustments in this case. 

\begin{figure}[!t]
\centering
\includegraphics[scale=0.40]{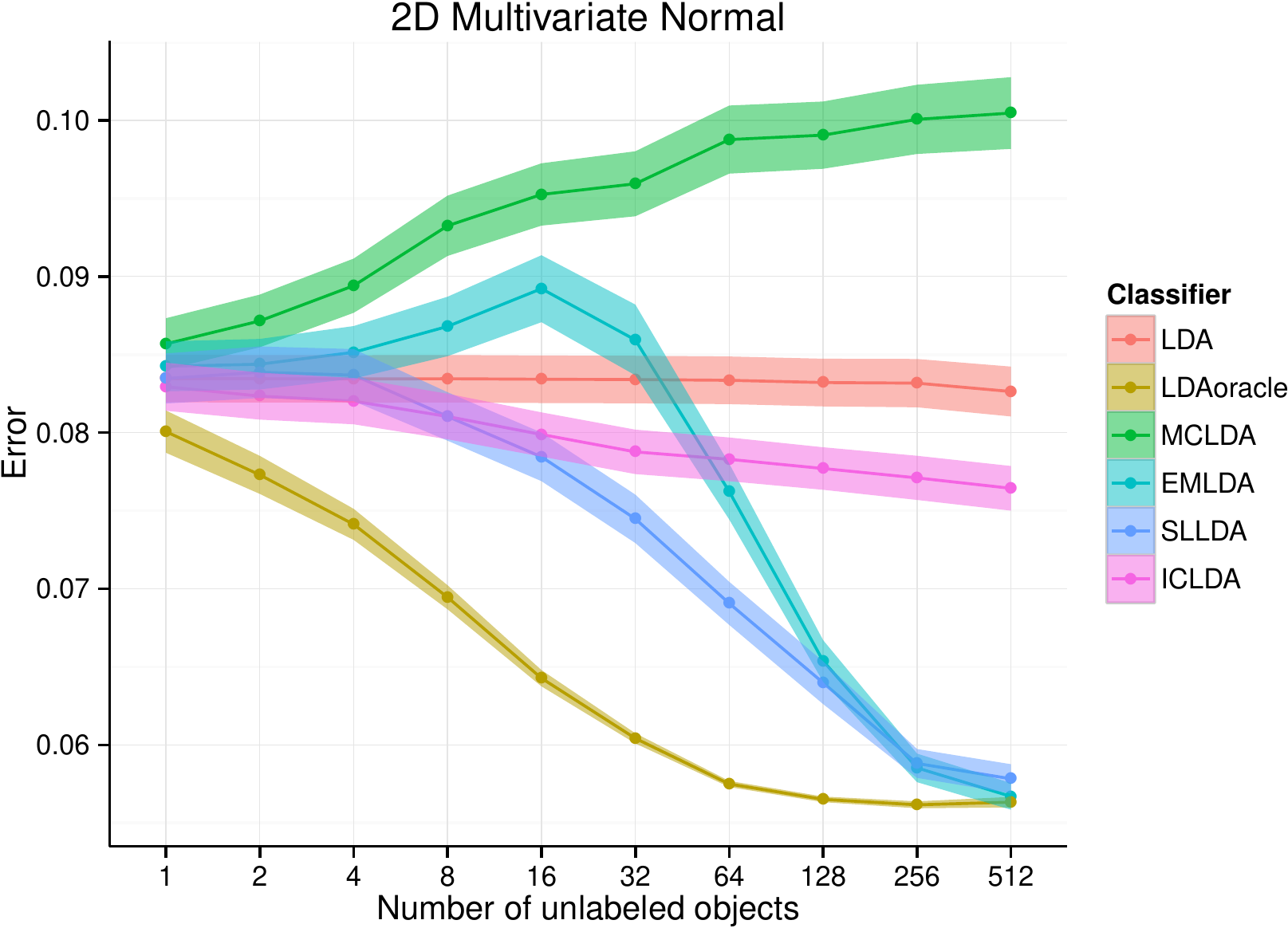}
\caption{Semi-supervised learning curve on the Gaussian data set using 500 repeats. The shaded regions indicate one standard error around the mean. Since their assumptions hold exactly, SLLDA and EMLDA work very well. ICLDA also outperforms the supervised LDA.}
\label{fig:learningcurvegauss}
\end{figure}

While one could claim that ICLDA is more robust, one could also expect ICLDA to never lead to any improvements. Figure \ref{fig:learningcurvegauss} shows the results when resampling from the data distribution in the second example and shows that ICLDA does lead to improvement on average in the second dataset, while not making the mistake in the first dataset where the LDA assumptions do not hold.

\subsection{Simulations on Benchmark datasets}
We test the behaviour of the considered procedures using datasets from the UCI machine learning repository \cite{Bache2013}, as well as from \cite{Chapelle2006}. Their characteristics can be found in Table \ref{table:datasets}.

\begin{table}[!t]
\caption{Description of the datasets used in the experiments}
\label{table:datasets}
\centering
\begin{tabular}{|rrrr|}
  \hline
 Name & \# Objects & \#Features & Source \\ 
  \hline
  Haberman & 305 &   4 & \cite{Bache2013} \\ 
  Ionosphere & 351 &  33 & \cite{Bache2013} \\ 
  Parkinsons & 195 &  20 & \cite{Bache2013} \\ 
  Pima & 768 &   9 & \cite{Bache2013} \\ 
  Sonar & 208 &  61 & \cite{Bache2013} \\ 
  SPECT & 265 &  23 & \cite{Bache2013} \\ 
  SPECTF & 265 &  45 & \cite{Bache2013} \\ 
  Transfusion & 748 &   4 & \cite{Bache2013} \\ 
  WDBC & 568 &  30 & \cite{Bache2013} \\
  BCI & 400 & 118 & \cite{Chapelle2006} \\ 
   \hline
\end{tabular}
\end{table}

A cross-validation experiment was carried out as follows. Each of the datasets were split into $10$ folds. Every fold was used as a validation set once, while the other nine folds were used for training. The data in these nine folds was randomly split into a labeled and an unlabeled part, where the labeled part had $\max(2d,10)$ objects, while the rest was used as unlabeled objects. This procedure was repeated $20$ times and the average error and average negative log likelihood (the loss function of LDA) on the test set was determined. The results can be found in Tables \ref{table:cvresults-error} and \ref{table:cvresults-loss}.

\begin{figure*}[!t]
\centering
\includegraphics[scale=0.5]{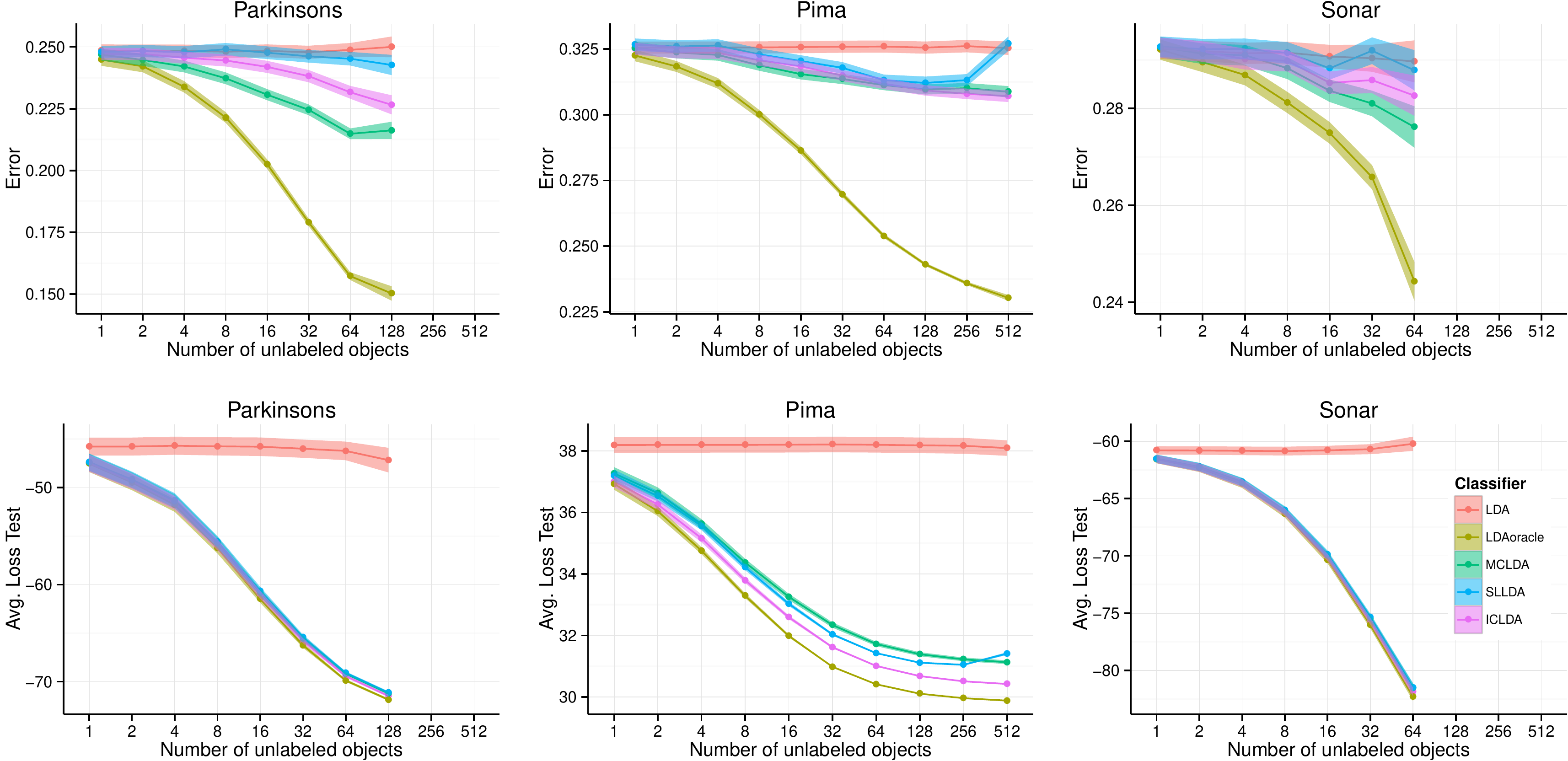}
\caption{Learning curves for increasing amounts of unlabeled data for the error rate as well as the loss (negative log likelihood) for three datasets using 500 repeats. The shaded regions indicate one standard error around the mean.}
\label{fig:errorcurves}
\end{figure*}

To study the behaviour of these classifiers for differing amount of unlabeled data, we estimated semi-supervised learning curves by randomly drawing $\max(2d,10)$ labeled objects from the datasets, and using an increasing randomly chosen set as unlabeled data. The remaining objects formed the test set. This procedure was repeated 500 times and the average and standard error of the classification error and negative log likelihood were determined. The learning curves for 3 datasets can be found in Figure \ref{fig:errorcurves}.

\begin{table*}
\caption{Average 10-fold cross-validation error and its standard deviation over 20 repeats. Indicated in $\mathbf{bold}$ is whether a semi-supervised classifier significantly outperform the supervised LDA classifier, as measured using a $t$-test with a $0.05$ significance level. \underline{Underlined} indicates whether a semi-supervised classifier is (significantly) best among the four semi-supervised classifiers considered.} \label{table:cvresults-error}
\centering
\begin{tabular}{|l|llllll|}
\hline
Dataset & LDA & LDAoracle & MCLDA & EMLDA & SLLDA & ICLDA \\ 
\hline
Haberman & $0.37 \pm 0.04$& $0.25 \pm 0.00$& $0.36 \pm 0.03$& $0.47 \pm 0.08$& $0.36 \pm 0.04$& $0.37 \pm 0.04$\\ 
Ionosphere & $0.21 \pm 0.02$& $0.15 \pm 0.01$& $\mathbf{\underline{0.18 \pm 0.02}} $& $0.57 \pm 0.04$& $0.20 \pm 0.02$& $\mathbf{0.18 \pm 0.01} $\\ 
Parkinsons & $0.27 \pm 0.03$& $0.15 \pm 0.01$& $\mathbf{\underline{0.22 \pm 0.03}} $& $0.41 \pm 0.05$& $0.26 \pm 0.03$& $\mathbf{0.23 \pm 0.03} $\\ 
Pima & $0.34 \pm 0.03$& $0.23 \pm 0.00$& $\mathbf{0.32 \pm 0.02} $& $0.37 \pm 0.03$& $0.35 \pm 0.02$& $\mathbf{\underline{0.31 \pm 0.02}} $\\ 
Sonar & $0.29 \pm 0.02$& $0.26 \pm 0.02$& $0.28 \pm 0.02$& $0.35 \pm 0.02$& $0.29 \pm 0.02$& $0.28 \pm 0.02$\\ 
SPECT & $0.31 \pm 0.03$& $0.18 \pm 0.01$& $\mathbf{\underline{0.25 \pm 0.02}} $& $0.62 \pm 0.03$& $0.33 \pm 0.03$& $0.30 \pm 0.03$\\ 
SPECTF & $0.32 \pm 0.03$& $0.24 \pm 0.01$& $\mathbf{\underline{0.28 \pm 0.03}} $& $\mathbf{0.28 \pm 0.05} $& $0.34 \pm 0.03$& $0.33 \pm 0.03$\\ 
Transfusion & $0.34 \pm 0.03$& $0.23 \pm 0.00$& $\mathbf{\underline{0.32 \pm 0.03}} $& $0.52 \pm 0.09$& $0.37 \pm 0.05$& $0.33 \pm 0.03$\\ 
WDBC & $0.11 \pm 0.01$& $0.04 \pm 0.00$& $\mathbf{0.09 \pm 0.01} $& $0.38 \pm 0.05$& $\mathbf{0.09 \pm 0.01} $& $\mathbf{\underline{0.08 \pm 0.01}} $\\ 
BCI & $0.21 \pm 0.01$& $0.16 \pm 0.01$& $\mathbf{\underline{0.20 \pm 0.01}} $& $0.21 \pm 0.02$& $0.21 \pm 0.02$& $\mathbf{0.20 \pm 0.01} $\\ 
\hline
\end{tabular}
\end{table*}

\begin{table*}
\caption{Average 10-fold cross-validation negative log-likelihood (loss) and its standard deviation over 20 repeats. Indicated in $\mathbf{bold}$ is whether a semi-supervised classifier significantly outperform the supervised LDA classifier, as measured using a $t$-test with a $0.05$ significance level. \underline{Underlined} indicates whether a semi-supervised classifier is (significantly) best among the four semi-supervised classifiers considered.} \label{table:cvresults-loss}
\centering
\begin{tabular}{|l|llllll|}
\hline
Dataset & LDA & LDAoracle & MCLDA & EMLDA & SLLDA & ICLDA \\ 
\hline
Haberman & $15.88 \pm 4.37$& $10.37 \pm 0.02$& $\mathbf{11.66 \pm 2.45} $& $\mathbf{12.02 \pm 0.35} $& $\mathbf{12.08 \pm 0.20} $& $\mathbf{\underline{10.89 \pm 0.16}} $\\ 
Ionosphere & $199.58 \pm 29.66$& $21.38 \pm 0.34$& $\mathbf{25.93 \pm 1.44} $& $\mathbf{22.55 \pm 0.40} $& $\mathbf{22.80 \pm 0.40} $& $\mathbf{\underline{22.22 \pm 0.33}} $\\ 
Parkinsons & $-40.76 \pm 11.11$& $-71.87 \pm 0.32$& $\mathbf{-71.05 \pm 0.40} $& $\mathbf{-71.12 \pm 0.40} $& $\mathbf{-71.03 \pm 0.38} $& $\mathbf{\underline{-71.44 \pm 0.31}} $\\ 
Pima & $41.98 \pm 2.99$& $29.88 \pm 0.02$& $\mathbf{31.74 \pm 0.99} $& $\mathbf{31.95 \pm 0.35} $& $\mathbf{32.07 \pm 0.36} $& $\mathbf{\underline{30.50 \pm 0.13}} $\\ 
Sonar & $-59.86 \pm 1.08$& $-83.05 \pm 0.59$& $\mathbf{-82.23 \pm 0.57} $& $\mathbf{-82.85 \pm 0.55} $& $\mathbf{-82.20 \pm 0.60} $& $\mathbf{-82.58 \pm 0.57} $\\ 
SPECT & $27.65 \pm 1.89$& $10.74 \pm 0.09$& $\mathbf{11.30 \pm 0.17} $& $\mathbf{12.63 \pm 0.18} $& $\mathbf{11.84 \pm 0.20} $& $\mathbf{\underline{11.19 \pm 0.13}} $\\ 
SPECTF & $178.42 \pm 2.48$& $148.13 \pm 0.68$& $\mathbf{148.78 \pm 0.69} $& $\mathbf{148.44 \pm 0.69} $& $\mathbf{149.18 \pm 0.72} $& $\mathbf{148.67 \pm 0.71} $\\ 
Transfusion & $17.00 \pm 2.61$& $11.48 \pm 0.02$& $\mathbf{12.23 \pm 0.54} $& $16.27 \pm 0.53$& $\mathbf{14.21 \pm 0.47} $& $\mathbf{\underline{11.88 \pm 0.17}} $\\ 
WDBC & $33.15 \pm 15.14$& $-28.06 \pm 1.29$& $\mathbf{-26.73 \pm 1.23} $& $\mathbf{-26.67 \pm 1.32} $& $\mathbf{-27.78 \pm 1.28} $& $\mathbf{\underline{-27.86 \pm 1.28}} $\\ 
BCI & $6.99 \pm 1.04$& $-21.04 \pm 0.41$& $\mathbf{-20.38 \pm 0.40} $& $\mathbf{-20.39 \pm 0.46} $& $\mathbf{-20.44 \pm 0.45} $& $\mathbf{\underline{-20.74 \pm 0.41}} $\\ 
\hline
\end{tabular}
\end{table*}

We find that overal in terms of error rates (Table \ref{table:cvresults-error}), MCLDA seems to perform best, being both more robust than the EM approaches as well as effective in using the unlabeled information in improving error rates. While ICLDA is robust and has the best performance on 2 of the datasets, it is conservative in that it does not show improvements in terms of the classification error for many datasets where the other classifiers do offer improvements.

The picture is markedly different when we consider the log likelihood criterion that supervised LDA optimizes, evaluated on the test set (Table \ref{table:cvresults-loss}). Here ICLDA outperforms all other methods on the majority of the datasets. 

\section{Discussion}
The results show that ICLDA provides the safest option among the semi-supervised approaches to LDA that we considered. At least in terms of the log-likelihood, it provides the best performance by far. A particularly interesting observation is that the implicit constraints, in many a case, seem to add a lot to the constraints that MCLDA enforces, as also the latter classifier is consistently outperformed by ICLDA in terms of the log likelihoods achieved on the test set. As yet, we have no full understanding in what way the implicit constraints add restrictions beyond the moment constraints, apart from the fact that the former are stricter than the latter.  A deeper insight into this issue might cast further light on the workings of ICLDA.

While it is the safest version, ICLDA may be \emph{too} safe, in that it does not attain performance improvement in terms of classification error in many cases where MCLDA or the EM approaches do offer improvements. In terms of the loss on the test set, however, ICLDA is the best performing method. Since this is the objective minimized by supervised LDA as well, perhaps this is the best we could hope for in a true semi-supervised adaptation of LDA. We found similar empirical and theoretical performance results in terms of improvements in the loss on the test set when applying the implicitly constrained framework to the least squares classifier \cite{Krijthe2013}. How then, this improvement in ``surrogate'' loss relates to the eventual goal of classification error, is unclear, especially for a non-margin based loss function such as the negative log likelihood \cite{Bartlett2006}. However, since ICLDA does offer the best behaviour of supervised LDA's loss on the test set, ICLDA could be considered a step towards a principled semi-supervised version of LDA.

An open question regarding the objective function associated with ICLDA is to what extent it is convex. The solution in terms of the responsibility vector $\mathbf{y}_u$ is non-unique: different labelings of the points can lead to the same parameters. In terms of the parameters, however, the optimization seems to converge to a unique global optimum. While we do not have a formal proof of this, as in the case of implicitly constrained least squares classification, we conjecture that the objective function is convex in the parameters, at least in the case in which we choose to parameterize LDA by means of its canonical parameters \cite{Lehmann1998}.  In this case, LDA does lead to a convex optimization problem.

We find that the behaviour of EMLDA is more erratic than that of SLLDA. The hard label assignments could have a regularizing effect on the semi-supervised solutions, making self-learning a good and fast alternative to self-learning. Note that safer versions of SLLDA and EMLDA could be obtained by introducing a weight parameter to control the influence of the unlabeled data \cite{McLachlan1975}. In the limited labeled data setting, it is hard to correctly set this parameter. While this may help when dealing with larger sample sizes, the constraint approaches bring us closer to methods that always perform at least as well as their supervised counterpart. 

\section{Conclusion}
ICLDA is a principled and robust adaptation of LDA to the semi-supervised setting. In terms of error rates, it may be overly conservative. When measured in terms of the loss on the test set, however, it outperforms other semi-supervised methods. It therefore seems that there are opportunities for robust semi-supervised learners, although the performance criterion that we should consider may not be the error rate, but rather the loss that the supervised learner minimizes \cite{Loog2014b}.


\section*{Acknowledgment}
This work was partly funded by project P24 of the Dutch public/private research network COMMIT.



\bibliographystyle{IEEEtranS}
\bibliography{IEEEabrv,library}
%



\end{document}